# Evaluating Synthetic Images as Effective Substitutes for Experimental Data in Surface Roughness Classification

[2]Binwei Chen, [1]Huachao Leng, [1]Chi Yeung Mang, [3]Tsz Wai Cheung, [2]Yanhua Chen, [2]Wai Keung Anthony Loh, [2]Chi Ho Wong, [1]Chak Yin Tang*

[1]*State Key Laboratory of Ultra-precision Machining Technology, Department of Industrial and Systems Engineering, The Hong Kong Polytechnic University, Hong Kong, China*

[2]*Division of Science, Engineering and Health Studies, The School of Professional Education and Executive Development, The Hong Kong Polytechnic University, Hong Kong, China*

[3]*Department of Science and Environmental Studies, The Education University of Hong Kong, Hong Kong, China*

*The first two authors contributed equally.*

*Email: cy.tang@polyu.edu.hk*

Abstract: Hard coatings play a critical role in industry, with ceramic materials offering outstanding hardness and thermal stability for applications that demand superior mechanical performance. However, deploying artificial intelligence (AI) for surface roughness classification is often constrained by the need for large labeled datasets and costly high-resolution imaging equipment. In this study, we explore the use of synthetic images, generated with Stable Diffusion XL, as an efficient alternative or supplement to experimentally acquired data for classifying ceramic surface roughness. We show that augmenting authentic datasets with generative images yields test accuracies comparable to those obtained using exclusively experimental images, demonstrating that synthetic images effectively reproduce the structural features necessary for classification. We further assess method robustness by systematically varying key training hyperparameters (epoch count, batch size, and learning rate), and identify configurations that preserve performance while reducing data requirements. Our results indicate that generative AI can substantially improve data efficiency and reliability in materials-image classification workflows, offering a practical route to lower experimental cost, accelerate model development, and expand AI applicability in materials engineering.

## 1. Introduction:

The growing demand for hard coatings across diverse industrial sectors has intensified the need for rigorous, reliable evaluation of surface roughness, which directly influences tribological behavior, fatigue life, adhesion, and mechanical performance [1]. Aluminum oxide ($Al_2O_3$) occupies an important role in this context: its combination of exceptional hardness, thermal stability, chemical inertness, and corrosion resistance makes it a widely adopted coating material for harsh-service components [2]. Beyond conventional protective roles, $Al_2O_3$ is considered for extreme environments such as near-solar applications, where it could act as a thermal shield against intense radiative heating and cyclical temperature extremes [3]. Its material properties also make it particularly amenable to advanced manufacturing since $Al_2O_3$ can be processed into intricate ceramic geometries via photocuring-based 3D printing techniques, enabling fabrications of complex and reproducible parts that are difficult to achieve with traditional methods. These favorable features support high-precision forming and tight process control.

Despite these advantages, accurate and scalable surface-roughness assessment remains challenging. High-quality surface characterization typically relies on high-resolution imaging, which can be time-consuming and costly when applied to large sample volumes or industrial inspection lines[4-8]. Preparing reliable images for AI training to support roughness classification is challenging, since modern classifiers typically require large, well-labeled datasets that capture the full range of surface morphologies, process variations, and defect modes to generalize reliably to new samples [6, 7]. For many laboratories and manufacturers, collecting and labeling sufficiently large experimental image sets at an ultra-high resolution is a limiting factor. Moreover, variations in imaging modality, lighting, and sample preparation introduce domain shifts that can erode classifier performance if not adequately represented in the training data.

Generative AI has emerged as a groundbreaking technology, revolutionizing the way we create and interpret visual content. By employing modern algorithms, such as those utilized in models like GAN, Stable Diffusion XL, generative AI is capable of producing high-quality synthetic images that closely resemble real-world visuals [9-13]. This capability has found applications across various domains, from artistic endeavors to scientific research, allowing users to generate diverse and compelling images with unprecedented ease and efficiency [14, 15]. In fields such as material engineering where sample size limitations often hinder comprehensive analysis of image classifications, synthetic images could serve as valuable substitutes, augmenting existing datasets and enabling more robust machine learning models [16-18]. By expanding the available training data, generative AI could improve classification accuracy and the reliability of image-recognition systems [19], enabling more efficient quality assessment without requiring ultra-high-resolution imaging, thereby reducing cost and challenges in measurements. Given the practical importance of $Al_2O_3$, we investigate whether synthetic $Al_2O_3$ images can reliably replace authentic images in downstream classification tasks without reducing performance on real test data [20, 21].

## 2. Methods

### 2.1. Sample fabrications

The 3D printing is performed on a lithography-based ceramic manufacturing (LCM) system (CeraFab CF7500, Lithoz GmbH, Austria) employing a digital light projection exposure mechanism with a wavelength of 460 nm. The process employed a commercial alumina slurry (LithaLox HP500, Lithoz GmbH, Austria), which contains high-purity alumina powder ($Al_2O_3$, 99.99%) dispersed in an acrylate-based photopolymer binder system. The slurry has a solid loading of 49 vol.%, a density of 2.49 g cm$^{-3}$, and a dynamic viscosity of 11.8 Pa·s at 20 °C and 3.6 Pa·s at 40 °C. The printing process is conducted with a layer thickness of 25 μm and an exposure energy of approximately 130 mJ cm$^{-2}$. The slurry temperature during printing is maintained at a range 20–25 °C. After printing, the green bodies are removed from the build platform and cleaned using LithaSol 20 solvent with an airbrush system. UV post-curing is performed for 5 min.

Thermal debinding of the printed green bodies is performed in a programmable furnace (FMJ-07/11, Facerom Intelligent Equipment Co., China). The temperature is increased from 25 °C to 75 °C over 4 hours and held for 18 hours, followed by heating to 115 °C over 4 hours with a dwell of 66 hours. The temperature is subsequently raised to 205 °C over 4 hours and held for 22 hours, then increased to 430 °C over 10 hours and further to 900 °C over 12 hours. The furnace is cooled to 600 °C over 8 hours and subsequently to 25 °C over 8 hours, resulting in a total debinding cycle of 156 hours. The brown bodies are sintered in a high-temperature furnace (FJL-09/16, Facerom Intelligent Equipment Co., China). The sintering profile is presented in Fig. 1b. The temperature is increased from 25 °C to 200 °C over 1 hour, followed by heating to 600 °C over 10 hours and 1150 °C over 6 hours. The temperature is then raised to 1600 °C over 9 hours and held for 2 hours. After sintering, the furnace is cooled to 1200 °C over 8 hours and subsequently to 50 °C over 12 hours, giving a total sintering cycle of 48 hours.

### 2.2: Authentic Image acquisition

Imaging is performed with an Olympus LEXT OLS5100 3D measuring laser scanning confocal microscope (LSCM) using a 20× lens and Smart Scan, yielding a 643 × 643 μm field of view. The material consisted of alumina additive-manufactured (AM) ceramics produced by Digital Light Processing (DLP) 3D printing. $Al_2O_3$ samples are labeled into three surface-roughness categories based on Sa: Low Roughness (Sa < 1 μm), Normal Roughness (1 μm ≤ Sa ≤ 3 μm), and High Roughness (Sa > 3 μm), where surface roughness is represented by color [5].

### 2.3: Synthetic Image acquisition

Synthetic images of $Al_2O_3$ are generated using Stable Diffusion XL through an image-to-image diffusion pipeline [9]. Authentic images are used as inputs to guide the generative process, enabling the synthetic images to preserve key visual and structural characteristics while introducing controlled variations. The generative procedure is applied separately to each quality level (High, Normal, and Low), producing synthetic counterparts corresponding to the authentic samples used in training. This enables systematic investigation of whether the generative images retain task-relevant structural information rather than merely replicating visual appearance.

All classification experiments are conducted using TensorFlow under Google Teachable Machine Platform. For each experimental setting, classification models are trained from scratch using the designated training sets and subsequently evaluated on the fixed authentic-image test set. In all experiments, the training sets are constructed from images numbered 1–50, consisting of authentic and/or synthetic images according to the experimental conditions, while the test sets consist exclusively of authentic images numbered 51–100. This design ensures that model evaluation is always performed on previously unseen authentic data. To evaluate the robustness of the classification framework, key training hyperparameters including epoch number, batch size, and learning rate are systematically examined. A baseline configuration of epoch = 100, batch size = 16, and learning rate = 0.001 is adopted. Each parameter is varied individually while keeping the others fixed. The examined hyperparameter ranges are summarized in Table 1.

**Table 1.** Hyperparameter ranges used in the optimization experiments.

| Hyperparameter | Baseline Value | Examined Values |
|---|---|---|
| Epoch | 100 | 5, 10, 20, 40, 100, 180 |
| Batch Size | 16 | 16, 32, 64, 128, 256, 512 |
| Learning Rate | 0.001 | $5\times10^{-5}, 1\times10^{-4}, 3\times10^{-4}, 1\times10^{-3}, 3\times10^{-3}, 1\times10^{-2}, 2\times10^{-2}$ |

To evaluate whether synthetic images can substitute for authentic training data, we design classification experiments under both binary and ternary settings. For the binary task, we frame a two-class problem comparing high and low roughness. We run two experimental conditions: (1) a control condition in which both classes are trained using only authentic images, and (2) a substitution condition in which one class (either High or Low) is represented in the training set by synthetic images while the other class used authentic images. Model performance is then measured on the authentic test sets to assess the impact of replacing real training images with generated ones. To further evaluate the robustness of synthetic image substitution in a more complex scenario, ternary classification experiments are conducted involving high, medium, and low roughness levels. A baseline configuration using entirely authentic images is compared with three substitution cases in which synthetic images replaced authentic images for different roughness levels. The detailed experimental configurations (B1–B2 for binary tasks and T1–T4 for ternary tasks) are summarized in Table 2.

**Table 2.** Experimental configurations for binary and ternary classification tasks. (B = Binary classification; T = Ternary classification; A = Authentic images; S = Synthetic images)

| Experiment | High | Medium | Low | Training Set | Test Set |
|---|---|---|---|---|---|
| B1 | A | – | A | 1–50 | A 51–100 |
| B2 | A | – | S | 1–50 | A 51–100 |
| T1 | A | A | A | 1–50 | A 51–100 |
| T2 | A | S | S | 1–50 | A 51–100 |
| T3 | S | A | S | 1–50 | A 51–100 |
| T4 | S | S | A | 1–50 | A 51–100 |

Model performance is evaluated using per-sample prediction correctness. A prediction is considered correct when the predicted surface roughness level (high, medium, or low) matched the ground-truth level of the test image.

The classification accuracy for each level is computed as:

$$Accuracy_i = \frac{N_{correct,i}}{N_{test,i}}$$

where $N_{correct,i}$ is the number of correctly classified samples for level i, and $N_{test,i}$ is the total number of test samples for that level (50 in this study). Although the training labels include image origin (authentic or synthetic), the evaluation metric considers only the roughness level, as the objective is to assess the performance of classification. To efficiently evaluate multiple trained models, a JavaScript-based evaluation script is developed to automatically load the exported Teachable Machine models via their URLs and perform batch inference on the test dataset. Classification accuracy is computed based on the results of prediction.

## 3. Results and Discussions

Representative LSCM surface topographies of the samples at three distinct roughness levels are presented in Figure 1. The color maps provide a visual qualitative assessment of the vertical deviations across the sample surfaces. Figure 1a shows a high-roughness material that exhibits significant topographic variation, characterized by large peak-to-valley transitions (i.e. the vertical change in height between the highest point (peak) and the lowest point (valley) within a specific surface) [22-24]. The upper region shows a dense cluster of high-altitude features (represented in orange and red), suggesting a fractured or deeply eroded surface with substantial material protrusion.

The medium roughness surface in Figure 1b displays a more grid-like texture. While the average height distribution is more uniform than in the high-roughness sample, there are distinct, localized "hot spots" (yellow/orange) and small pits (deep blue) distributed across the scan area. In contrast, the low-roughness sample in Figure 1c shows the most homogeneous surface profile. The dominance of uniform green hues indicates a narrow height distribution with minimal vertical deviation. The absence of significant peaks or deep valleys suggests a highly polished, specular surface finish characteristic of fine abrasive processing.

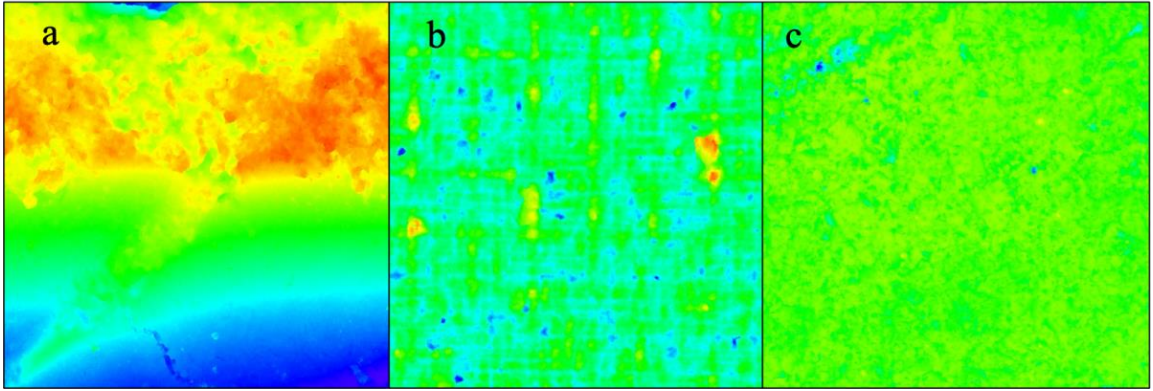

**Figure 1.** Representative LSCM surface images of $Al_2O_3$ samples showing different surface roughness levels: (a) high, (b) medium, and (c) low. The transition from (a) to (c) illustrates a clear reduction in surface complexity and height amplitude, moving from a stochastic, highly textured topography to a nearly planar surface.

Figure 2 presents a side-by-side comparison between authentic LSCM images (top row) and synthetic images (bottom row) across three distinct surface roughness regimes for $Al_2O_3$. Although the color edges in the synthetic images are not as sharp as those in the authentic image, the visual correlation between the two rows demonstrates the high fidelity of the synthetic generation process [25, 26]. The synthetic image (Figure 2b) accurately replicates the complex, stochastic morphology seen in the authentic sample (Figure 2a). It successfully captures the large-scale vertical deviations, including the prominent central peak (orange/red) and the deep peripheral valleys (blue), maintaining the irregular "island-like" features characteristic of high-roughness surfaces. For the medium roughness level, the synthetic model also effectively reproduces the grid-like texture observed in the authentic scan (Figure 2c vs Figure 2d). The spatial distribution of localized peaks and pits is almost indistinguishable, indicating that the generation process can mimic the structured patterns typical of mechanical finishing. In the low-roughness samples (Figure 2e vs Figure 2f), both authentic and synthetic images exhibit a highly uniform topography dominated by a narrow height distribution (green). The synthetic version captures the subtle, fine-grained texture of a polished surface without introducing obvious noise or artifacts. Hence, the structural similarity between the authentic and synthetic pairs across all three levels suggests that the synthetic data is representative of the actual $Al_2O_3$ surface characteristics. This indicates that the synthetic images could potentially be used to augment datasets for training predictive models or for virtual surface analysis of $Al_2O_3$ materials [27, 28].

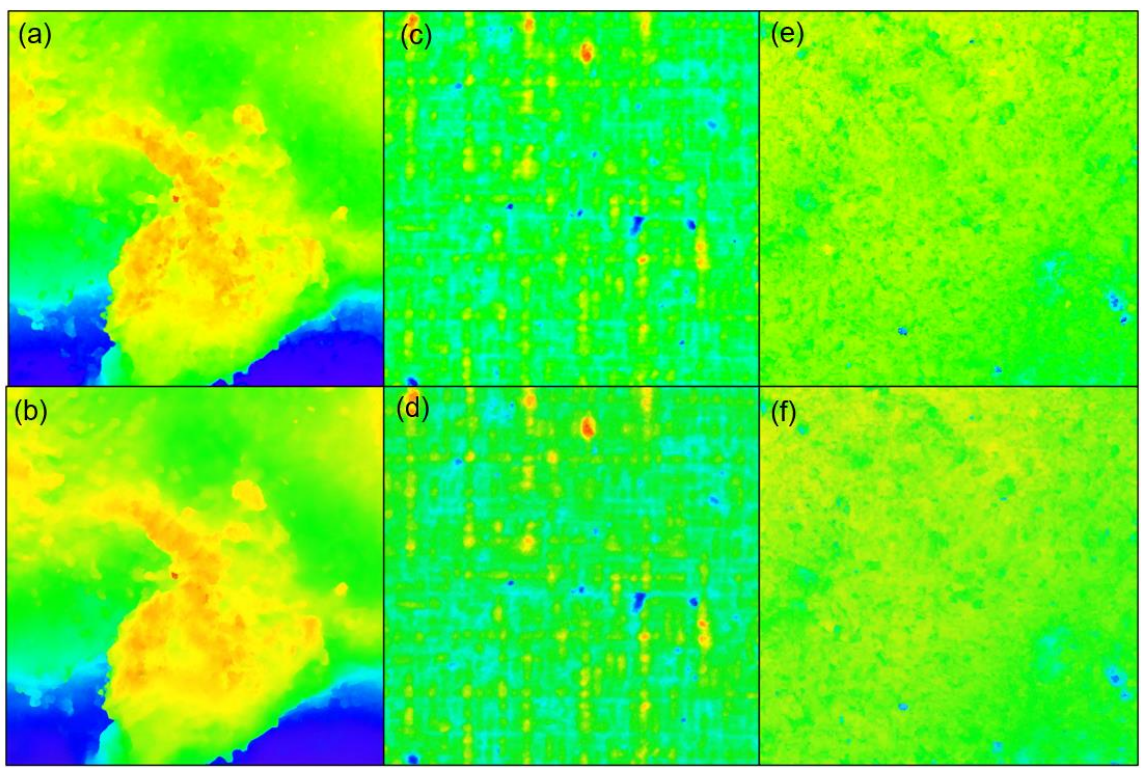

**Figure 2.** Comparison between authentic and synthetic LSCM images of $Al_2O_3$ samples at three surface roughness levels. The top row (a, c, e) shows authentic images, and the bottom row (b, d, f) shows synthetic images, corresponding to high, medium, and low roughness from left to right. For clarity, we display a single representative image contrast for each category.

A binary classification experiment is conducted to determine whether synthetic images can successfully replace authentic training images without degrading classifier performance. Two experimental configurations (B1 and B2 in Table 2) are evaluated: each configuration substituted synthetic images for one roughness level in the training set while retaining authentic images for the other level. Both configurations are trained and tested using the same model architecture and hyperparameter settings, and performance is measured on the held-out authentic test set to ensure assessment of real-world generalization. These outcomes indicate three important points. First, synthetic images closely capture the discriminative surface features needed by the classifier to distinguish high from low roughness, since replacing one class's real images with synthetic ones does not reduce test accuracy. Second, the synthetic data do not introduce systematic biases or spurious artifacts that would harm generalization to authentic images (i.e. test performance remains essentially unchanged). Third, the slight blur present in some training images does not affect classification accuracy, so acquiring ultrahigh-resolution images with expensive equipment may not be necessary, provided the generated images are of sufficient quality [29].

To further examine whether this observation remains valid in a more complex scenario, a three-level classification experiment is conducted. The three-level classification results corresponding to T1–T4 in Table 2 are presented in this section. Across all four experimental configurations, classification performance is good in T1, where the model achieved 98% accuracy on High-roughness images and 100% accuracy on both Normal and Low-roughness images, with only negligible variation between experiments [30].

Hyperparameter optimization is performed for the three substitution configurations T2–T4, while the authentic-only configuration T1 served as the baseline reference. The influence of training hyperparameters on classification accuracy is summarized in Figure 3. As shown in Figure 3(a), the classification accuracy remains relatively stable across different epoch numbers. Increasing the number of epochs beyond moderate values does not lead to noticeable performance improvement. Figure 3(b) illustrates the effect of batch size. The model maintains consistent accuracy over a wide range of batch sizes, indicating that the classification framework is not sensitive to this parameter within the tested range. In contrast, the learning rate exhibits a more pronounced influence on model performance. As shown in Figure 3(c), moderate learning rates produce stable classification accuracy, whereas excessively large learning rates lead to a clear decrease in performance.

Since additional epochs and large batch-size tuning do not substantially improve accuracy, training can be kept efficient (fewer epochs, flexible batch sizes) without sacrificing performance. Similar hyperparameter trends across authentic-only and substitution configurations indicate the stability of framework when synthetic images are included, where synthetic data do not introduce unusual sensitivity to training settings. Compared with the binary classification results, the consistent performance in the three-level task further confirms that replacing authentic training images with synthetic images does not significantly affect accuracy on authentic test data. This suggests that verifying the surface roughness of $Al_2O_3$ may not require ultra-high-resolution imaging systems, provided that the synthetic images are sufficiently realistic.

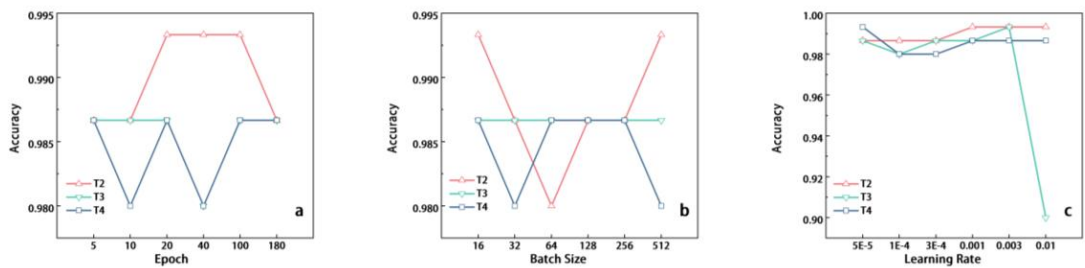

**Figure 3.** Hyperparameter optimization results for the substitution experiments (T2–T4). (a) Influence of epoch number on classification accuracy. (b) Influence of batch size on classification accuracy. (c) Influence of learning rate on classification accuracy.

Despite the encouraging results, several limitations should be acknowledged. First, the present study focuses on a single material system ($Al_2O_3$) and a specific surface roughness classification task. The generalizability of synthetic image substitution across different materials and imaging conditions remains to be further investigated. Second, the current evaluation primarily relies on classification accuracy on authentic test images. Future studies may incorporate additional evaluation metrics and more diverse datasets to better assess the robustness of synthetic substitution in materials image analysis. Furthermore, exploring more fine-grained classification tasks may help clarify the limits of replacing experimental images with synthetic data.

## 4. Conclusions

This study investigated the feasibility of using AI-generated synthetic images as substitutes for authentic experimental data in image classification tasks, using $Al_2O_3$ surface images obtained by laser scanning confocal microscopy (LSCM) as a representative material system. Through a series of progressive experiments, replacing authentic training images with synthetic ones resulted in classification performance comparable to that of authentic-only training. These results suggest that reliable image-based classification does not necessarily require large volumes of experimentally acquired images. Instead, synthetic images can effectively complement or partially substitute authentic datasets, thereby reducing experimental imaging demands and improving data efficiency in materials characterization workflows.

**Conflict of interests**

The authors declare no conflict of interest

**Author Contributions**

Conceptualization, C.Y.T, C.H.W.; methodology, C.Y.T, C.H.W, Y.C.; validation, B.C, H.L, C.Y.M.; formal analysis, C.Y.T, C.H.W, B.C, H.L, C.Y.M.; data curation, B.C.; writing, C.H.W, B.C, T.W.C, C.Y.T.; visualization, C.H.W., W.K.A.L, C.Y.T.; Resource, C.Y.T

**Data Availability Statement**

Data availability is possible upon reasonable request.

**Acknowledgments**

The work described in this paper was mainly supported by the funding support to the State Key Laboratories in Hong Kong from the Innovation and Technology Commission (ITC of the Government of the Hong Kong Special Administrative Region (HKSAR) of China, and The Hong Kong Polytechnic University (Project Code: 1-BBTN) and CPCE Research Fund (SEHS-2024-354(I)) at the Hong Kong Polytechnic University.